%% file: main.tex
\documentclass[english]{article}

\usepackage{microtype}
\usepackage{graphicx}
\usepackage{subfigure}
\usepackage{makecell}
\usepackage{booktabs} 
\usepackage{multirow}

\usepackage{hyperref}
\usepackage[nonatbib,preprint]{mystyle}

\usepackage{qiangstyle}

\title{Centroid Transformers: Learning to Abstract with Attention}

\author{%
  Lemeng Wu \thanks{Equal Contribution} \\
  UT Austin \\
  \texttt{lmwu@cs.utexas.edu} \\
    \and
   Xingchao Liu \textsuperscript{*} \\
  UT Austin\\
  \texttt{xcliu@utexas.edu} \\
  \and
   Qiang Liu \\
  UT Austin \\
  \texttt{lqiang@cs.utexas.edu} \\
}
\date{}

\begin{document}

\twocolumn[

    \maketitle

]
{
  \renewcommand{\thefootnote}%
    {\fnsymbol{footnote}}
  \footnotetext[1]{Equal contribution}
}



\begin{abstract}
Self-attention, as the key block of transformers,
is a powerful mechanism for extracting features from the inputs.
In essence,
what self-attention does is to infer the
\emph{pairwise  relations} between  the  elements of the inputs,
and 
modify the inputs by propagating information between the input pairs.
As a result, it maps $N$ inputs to $N$ outputs
and casts a quadratic $O(N^2)$
memory and time complexity.
%
We propose \emph{centroid attention},
a generalization of self-attention that maps $N$ inputs to $M$ outputs ($M\leq N$),
such that the key information in the inputs is summarized in the smaller number of outputs (called centroids).
We design centroid attention by amortizing
the gradient descent update rule of a clustering objective function on the inputs, which reveals a underlying connection between attention and clustering.
By compressing the inputs to the centroids,
we extract the key information useful for prediction and also reduce the computation of the  attention module and the subsequent layers.
We apply our method to various applications, including abstractive text summarization, 3D vision, and image processing.
Empirical results demonstrate the effectiveness of our method over the standard transformers.

\end{abstract}

\section{Introduction}
\input{text/introduction}

\section{Centroid Transformer}
\input{text/method}
\section{Experiments}
\input{text/experiments}

\section{Related Works}
\input{text/related_works}

\bibliography{main}
\bibliographystyle{b2021}

\newpage\clearpage
\appendix
\onecolumn
\input{text/appendix_vis}
\end{document}

%% file: text/introduction.tex
Recently, transformer \citep{vaswani2017attention} has emerged to be
one of the most important neural architectures 
and has achieved remarkable successes on various tasks such as 
language modeling~\citep{irie2019language, jiao2020tinybert}, 
 machine translation~\citep{vaswani2017attention,zhang2018improving,wang2019learning},   computer vision~\citep{carion2020end, dosovitskiy2020image}, and many others. 

What makes transformers unique is the extensive usage of the self-attention mechanism~\citep{vaswani2017attention}. A self-attention block is placed in each stage of the transformer to gather information \emph{globally} from the input sequence. 
A self-attention module takes in $N$ inputs, and returns $N$ outputs of the same size.
For each element in the input, it assigns an attention weight to every other element in the input to find out who it should pay more attention to,  
and perform a weighted sum to aggregate the information from the relevant inputs. 

Intuitively, 
the self-attention modules 
can be viewed as conducting \emph{interactive reasoning}, 
inferring the pairwise interacting relations between the elements of inputs and 
propagating information between pairs of elements. 
Naturally, a key drawback of the pairwise interaction  is that it casts an $O(N^2)$ memory and time complexity, where $N$ is the number of input elements, making it a major computational bottleneck in transformers. 
This necessitates an emerging line of research on approximating self-attention modules to gain higher computational efficiency 
\citep{kitaev2019reformer, katharopoulos2020transformers, wang2020linformer}. 


In this work, we develop a variant of attention module  
for conducting \emph{summative reasoning} rather than \emph{interactive reasoning}. 
Our goal is to take $N$ input elements and return a smaller number $M$ of outputs ($M\leq N$), such that the key information in the $N$ inputs are summarized in the $M$ outputs. 
If $N=M$, our new module  reduces to standard self-attention  (except an extra skip connection link). 
However, by setting $M < N$, 
we creates an information bottleneck and enforce the network to filter out the useless information (i.e., dimension reduction), and also 
 improve the computational cost from $O(N^2)$ to $O(NM)$.  
 In addition, once the number of inputs is reduced, the complexity of the subsequent 
 modules/layers is also reduced accordingly (e.g., applying self-attention on the output will yield a $O(M^2)$ complexity). 

Intuitively, we hope that the $M$ output elements are representative of the $N$ inputs. 
This can be viewed as a clustering process in which we
find $M$ ``centroids'' for the inputs.
Our new module, which we call \emph{centroid attention layer}, 
is obtained by \emph{``amortizing'' the gradient descent update rule on a clustering objective function}.  
It exploits a novel connection between
self-attention and clustering algorithms: 
we 
write down a clustering objective function on the input data under a trainable similarity metric, and derive its gradient descent update rule; 
we then observe that the gradient descent update yields a generalization of self-attention layer and use it to motivate the design of our new module. 

Using our new modules, we build \emph{centroid transformers},  
in which we insert our centroid attention modules between typical self-attention modules. 
%
We apply centroid transformers on several challenging scenarios, ranging from natural language processing to computer vision. On abstractive text summarization, centroid transformer beats the vanilla transformer with about only 50\% computational cost. On 3D point cloud classification and image classification tasks, centroid transformer achieves substantially higher accuracy as well as computational efficiency compared with the state-of-the-art transformer networks. 
We also use centroid attention to 
replace the dynamic routing module in the 3D Capsule Network \citep{zhao20193d} for point cloud reconstruction, 
which we find yields lower construction error,
reduced storage consumption, and more semantically meaningful latent representation. 

%% file: text/method.tex
\begin{figure}[t]
    \centering
    \includegraphics[width = 0.48\textwidth]{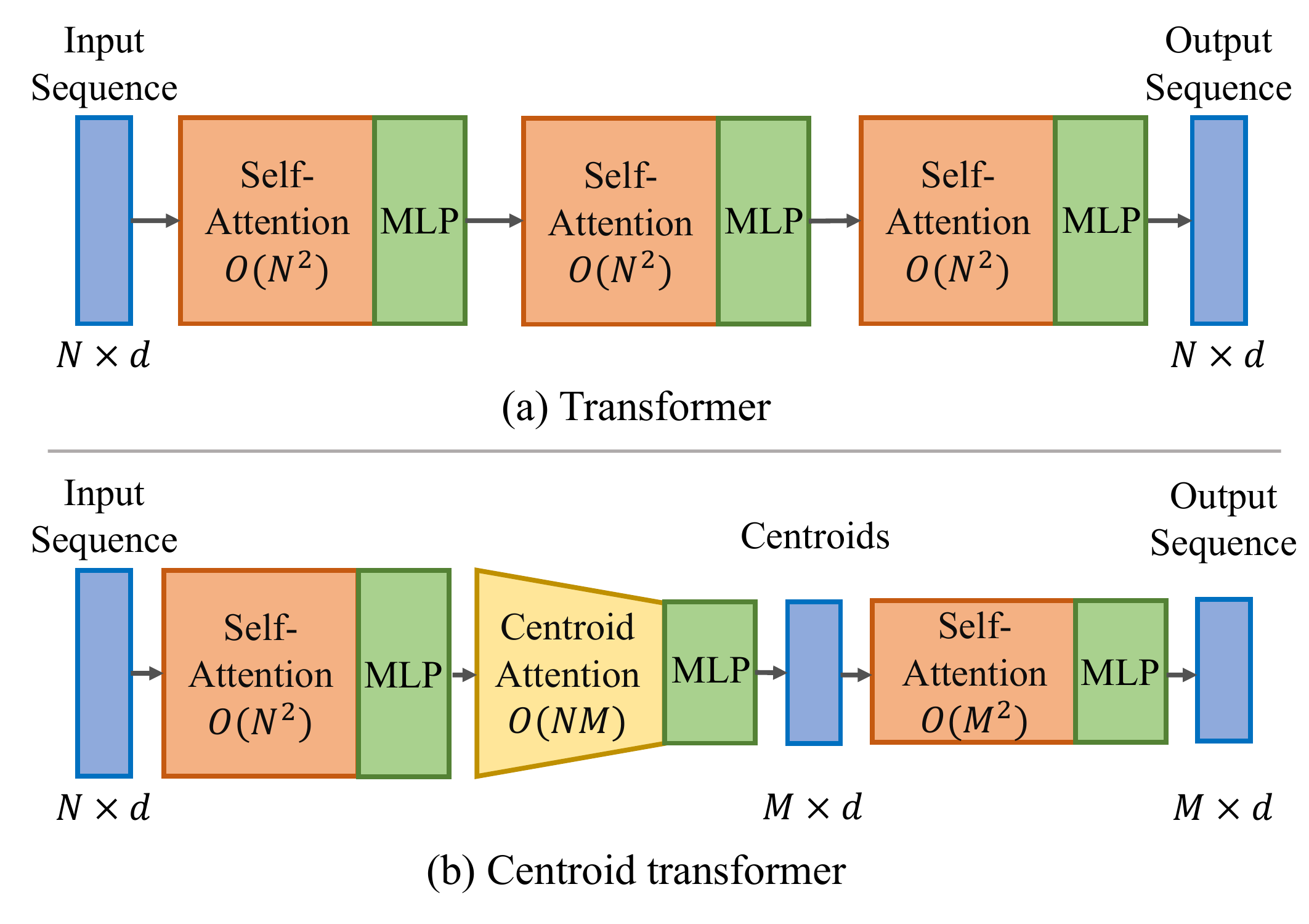}
    \vspace{-20pt}
    \caption{The vanilla transformer  (a)
    which maps $N$ inputs to $N$ outputs; 
    and our centroid transformer (b) which summarizes $N$ inputs into $M$ ``centroid'' outputs ($M\leq N$) to save computational cost and filter out useless information simultaneously. 
    }
    \label{fig:intro}
\end{figure}

We first introduce the standard self-attention mechanism used in vanilla transformers in Section~\ref{sec:selfattention}, 
and then 
derive the centroid attention mechanism by drawing inspiration from the gradient descent update of a clustering objective function 
and discuss the related centroid transformer in Section~\ref{sec:centroid}.

\subsection{Self-Attention Mechanism}\label{sec:selfattention}

Let $\{\bm x_i\}_{i=1}^N \in \R^{N \times d}$ be a set of input vectors, where we may view each vector $\bm x_i$ as a data point or ``particle".
Each $\bm x_i$ can be a word embedding vector in natural language processing, an image patch in computer vision, or a point in 3D point cloud. 
A self-attention 
module 
can be viewed as updating $\{\bm x_i\}$ in parallel via 
\begin{equation}
\label{eq:trans_attn}
    \bm{x}_i'~\leftarrow~\bm{x}_i + \epsilon \sum_{j=1}^N 
    \sum_{\ell=1}^L
    \text{sim}_\ell  \left(\bm{x}_i, \bm{x}_j\right)\times v_\ell (\bm{x}_j), ~~ \forall i \in [N]
\end{equation}
Here, each $\text{sim}_\ell(\cdot, \cdot)$ is a similarity function and $v_\ell(\bm x_i) \in \RR^d$ is a value function; each $\ell$ is considered to be a head in the attention and process a specific aspect of the inputs; $\epsilon$ is a positive constant.

Intuitively, 
the self-attention module 
evaluates the similarity (or attention score) between pairs of input vectors,  and updates each vector with the sum of the inputs weighted by the similarity scores.  
In practice, the similarity function is often defined as 
\begin{equation}
\label{eq:softmax}
\text{sim}_\ell \left(\bm{x}_i, \bm{x}_j\right)=\frac{\exp(Q_\ell^\top(\bm{x}_i)K_\ell(\bm{x}_j))}{\sum_{k=1}^N \exp (Q_\ell^\top(\bm{x}_i)K_\ell(\bm{x}_k))},
\end{equation}
where $Q_\ell(\bm x)$ and $K_\ell(\bm x)$ are the ``query'' and ``key'' functions of the $\ell$-th head. 
Eq.~\eqref{eq:trans_attn} and~\eqref{eq:softmax} illustrate how the features are processed by one head. 
 A complete transformer encoder with $T$ layers is the composition of self-attention layers and multi-layer perceptrons (MLPs); see Figure~\ref{fig:intro}. 
 Obviously, the time and memory complexities 
 of self-attention are quadratic on $N$, i.e., $O(N^2)$, which form the key computational bottleneck when applying self-attention on long sequences and large batches of data. 
 %

%

\subsection{Centroid Attention} 
\label{sec:centroid}
Self-attention updates the inputs $\{\bm x_i\}$ based on their interacting relations, 
obtaining an output of the same size. 
In this work, 
we propose a mechanism that maps the $N$ inputs $\{\bm x_i\}_{i=1}^n$ to $M$ output vectors $\{\bm u_i\}_{i=1}^M, M\leq N$, such that each $\bm u_i \in \RR^d$ can be viewed as a centroid of the inputs $\{\bm x_i\}_{i=1}^N$. Namely, we hope the module to be able to effectively perform a  clustering-like operation on the inputs, where $\{\bm u_i\}_{i=1}^M$  is a compression of $\{\bm x_i\}_{i=1}^M$ while inheriting the key information. 

%
%
%
%
%

Let $\{\phi_\ell(\vv x_i, \vv u_j)\colon~\forall \ell\in [L]\}$ be a set of measures of similarity between centroid $\vv u_j$ and input $\vv x_i$. 
Ideally, we may want to construct the centroids by optimizing the following 
general ``soft K-means'' 
objective function for clustering,
\begin{equation}
\label{soft-kmeans}
    \{\bm{u}^*_j\} = \argmax_{\{\bm{u}_j\}} \sum_{\ell=1}^L \sum_{i=1}^N \frac{1}{\alpha} \log \left( \sum_{j=1}^M \exp \left(\alpha \phi_\ell( \bm{x}_i, \bm{u}_j) \right) \right).
\end{equation}
Here, $\alpha > 0$ is a positive coefficient. 
If $\alpha \to +\infty$ and $L=1$, then the objective reduces to $\sum_i \max_j \phi_1(\vv x_i, \vv u_j)$, which coincides with the typical k-means objective function when $-\phi_1(\cdot,\cdot)$ is a distance measure. 
In general, the objective 
\eqref{soft-kmeans} 
obtains centroids to represent the inputs 
based on multiple similarity functions $\{\phi_\ell\}$.

\begin{figure}[t]
    \centering
    \includegraphics[width = 0.48\textwidth]{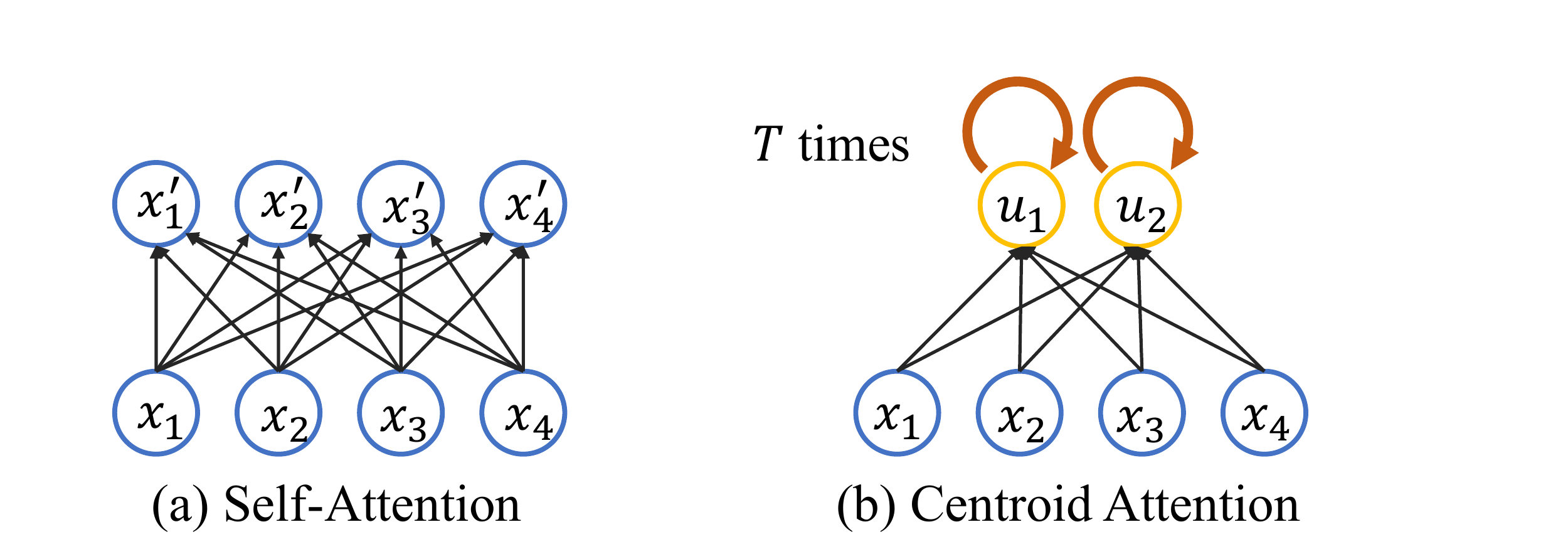}
    \caption{(a) The self-attention module, 
    which  modifies the input sequence $\{\vv x_i\}$ into $\{\vv x_i'\}$ by updating them  with pairwise interactions (see Eq~\eqref{eq:trans_attn}). (b) The centroid attention module,  which transforms the input sequence $\{\vv x_i\}$ into a set of centroids $\{\vv u_i\}$ by  first initializing the centroids and then updating them by interacting with the elements in the inputs (see Eq~\eqref{equ:ourmethod}).  
    }    
    \label{fig:abstract} 
\end{figure}

By solving the optimization  
 with gradient descent, we can unroll \eqref{soft-kmeans} into an iterative update of form: 
 \begin{align}
  &\text{Initialization:}~~~\{\vv u_j^0\}_{j=1}^M = I(\{\vv x_i\}_{i=1}^N) \notag \\
  &\text{For $t=1,2,\dots, T$,} \notag\\ 
   &~~~~~    \bm{u}_j^{t+1} \leftarrow \bm{u}_j^t + \epsilon \sum_{\ell=1}^L \sum_{i=1}^N \mathrm{sim}_\ell( \bm{x}_i, \bm{u}_j^t) V_\ell( \vv x_i, ~\vv u_j^t),\label{equ:ourmethod}
 \end{align}
 where $I(\cdot)$ denotes a mechanism for initializing the centroids and each of the following $T$ steps conducts gradient descent of \eqref{soft-kmeans}, with
 \begin{align*} 
 & \mathrm{sim}_\ell(\vv x_i, \vv u_j) = \frac{\exp(\alpha \phi_\ell(\vv x_i, \vv u_j))}{\sum_{k=1}^M \exp (\alpha \phi_\ell(\bm{x}_i, \bm{u}_k))}, \\
 &  V_\ell( \vv x_i, ~\vv u_j) = \nabla_{\bm{u}_j} \phi_\ell(\vv x_i, \vv u_j).
 \end{align*}
 Clearly, the gradient update above can be interpreted as a multi-head attention mechanism, 
 with $\mathrm{sim}_\ell(\cdot,\cdot)$ and $V_\ell(\cdot,\cdot)$ being the similarity function and the value function of the $\ell$-th head. 
 The initialization $I(\cdot)$ together with the $T$ steps of attention like updates in \eqref{equ:ourmethod} above form a \emph{centroid attention module}. 
 See Figure~\ref{fig:abstract} for an illustration.

 In practice, we find it works well to 
  set $\epsilon = 1 / T$ by default. Moreover, 
  in settings when computational efficiency is important, 
  we set $T = 1$ for a good performance-efficiency trade off. 
  The initialization step can vary in different tasks: 
  we can, for example, draw 
  $\{\vv u_j^0\}_{j=1}^M$  from $\{\vv x_i\}_{i=1}^N$
  by random sampling without replacement or farthest point sampling; we can also directly define $I(\cdot)$ to be a trainable fully connected or convolution layer. 
  See more discussion in the experiment section.

 Although 
 both $\mathrm{sim}_\ell(\cdot,\cdot)$ and $V_\ell(\cdot,\cdot)$ are determined by $\phi_\ell$ following the derivation above. In practice, we can define them separately in more flexible forms based on practical needs. 
 For example, we may define  $\mathrm{sim}_\ell(\cdot,\cdot)$  by setting 
 $\phi(\vv x_i, \vv u_j) = Q_\ell^\top(\bm{u}_j) K(\bm{x}_i)$  as typical self-attention, while setting $V_\ell( \vv x_i, ~\vv u_j)$ with a separate trainable value function. 
 



Our module includes  self-attention as a special case when we  \emph{i}) set $M = N$, \emph{ii}) initialize   $\{\vv u_i\}$ to be the same as $\vv x_i$, and \emph{iii}) iterate for one step ($T=1$). 
Therefore, our derivation also reveals a  close connection between gradient-based clustering and self attention. Note that~\citet{ramsauer2020hopfield} discusses a connection between Hopfield update rule and self-attention, 
which is related but different from our perspective. 


\paragraph{KNN Approximation}
The computational cost of the attention is $O(NM)$, which can still be expensive if $N $ and $M$ are large, 
To further save the computational cost,  we apply 
a K-nearest neighbor (KNN) approximation to \eqref{equ:ourmethod}, yielding 
$$
\bm{u}_j^{t+1} \leftarrow \bm{u}_j^t + \epsilon \sum_{\ell=1}^L \sum_{i\in \mathcal N_{j,k}} \mathrm{sim}_\ell( \bm{x}_i, \bm{u}_j^t) V_\ell( \vv x_i, ~\vv u_j^t),
$$
where $\mathcal N_{j,k}$ denotes the  K-nearest neighbor of $\vv u_j$ inside $\{\vx_i\}_{i=1}^N$, that is, 
$\mathcal N_{j,k}  = \{i_{(1)},\ldots, i_{(k)}\}$, where $\{i_{(1)},\ldots, i_{(n)}\}$ is the ascending reordering of $\{1,\dots, n\}$ according to some distance metric $d(\vx_i, \vv u_j)$. 
In practice, we define the distance by $
d(\vx_i, \vv u_j)= || \vx_i - \vv u_j||_2$. 

As shown in our experiments, the KNN approximation is particularly useful for the point cloud classification task, 
in which the length of the inputs elements is 
$N=1024$, which is beyond the capacity of our GPU memory even  for a single data point (i.e., a batch size of 1). 

\paragraph{Centroid Transformer} 
As shown in Fig.~\ref{fig:intro},
 we construct a centroid transformer architecture by alternating between typical self-attention blocks and centroid attention blocks. 
 This allows us to gradually abstract the input sequence to increasingly short representations, which filters out useless  information and simultaneously saves the computation cost. 
 As vanilla transformer, we insert 
fully connected  MLP layers 
with residual links
between the attention modules.   

%% file: text/experiments.tex

We apply our centroid attention block
on different kind of tasks with transformer or capsule structures to show our powerful abstract ability as well as the computational efficiency. We show that our centroid transformer structure has a strong performance in various of tasks including text summarization, point cloud classification, point cloud reconstruction and image classification. On all the tasks, we outperform baseline transformer with lower computational and storage cost.

\begin{table*}[t]
    \centering
   \begin{tabular}{l|ccccc}
        \hline
        Models & MACs(M) & MBS/GPU & ROUGE-1 & ROUGE-2 & ROUGE-L\\
        \hline
        Transformer & 523.2 & 192 & 32.987 & 15.286 & 30.771\\
        Ours-Random  & \textbf{262.9} & \textbf{230} & 30.310 & 12.752 & 27.823 \\
        \textbf{Ours-MP}  & \textbf{262.9} & \textbf{230} & \textbf{34.651} & \textbf{16.468} & \textbf{32.415}\\ 
        \hline
    \end{tabular}
    \caption{Results on Gigaword text summarization task  (MBS=Maximal Batch Size, MP = Mean-Pooling). The MACs (Multiply-add ACcumulation) is only computed for the encoder, assuming the length of sequence is 45 (the maximal length of sequence in the dataset). Though centroid transformer with random initialization (Ours-Random) performs worse than the baseline, centroid transformer with mean-pooling being the initialization method (Ours-MP) yields the best ROUGE score (See~\citep{lin2004rouge} for its definition) with $~$50\% computational cost compared to the original transformer.
    }
    \label{tab:nlp}
 
\end{table*}

\subsection{Abstractive Text Summarization}

We test centroid transformer on
abstractive text summarization,
a classic task in natural language  processing.
The task is to provide a short summary text (in several words) for a paragraph or a long sentence. 

\textbf{Data}  We use the annotated Gigaword corpus~\citep{rush2015neural} as our benchmark to compare different methods. The dataset is pre-processed with the tools provided by the authors. The corpus contains about 3.8 millions of training examples. The script also has 400,000 extra examples for validation and test. We randomly sample 2,000 examples for validation and test respectively, as in ~\cite{nallapati2016abstractive}. All the models are validated and tested with the same validation and test set. The performance of the generated summaries is measured by Recall-Oriented Understudy for Gisting Evaluation (ROUGE) ~\citep{lin2004rouge}. ROUGE-1/ ROUGE-2/ROUGE-L evaluates the quality of unigrams/bigrams/whole sentence in the generated summary.

\textbf{Model Configuration} \space We construct our centroid transformer by replacing the second self-attention module in the baseline transformer encoder with our centroid attention module.

$M$ is set to $N/3$ so that
our centroid attention module compresses $N$ input points into $N/3$ centroids.
The rest of the parts are kept unchanged. When decoding, the cross-attention is applied between the generated sequence and the centroids.
We test two initialization schemes for the centroid attention, random sampling and mean pooling. Random sampling means we randomly sample $N/3$ elements from the original sequence as the initial centroids, while mean pooling refers to apply mean pooling on every three elements.

Our baseline transformer follows the original encoder-decoder structure in ~\cite{vaswani2017attention}, with 4 layers in both encoder and decoder. We use a word embedding with 512 dimensions. All the models are trained for 30 epochs with Adam optimizer and the same learning rate. The number of warm-up steps is set to 4,000. At decode time, beam search with size 10 is exploited to generate the summary. Both self-attention and cross-attention is enabled in the decoder.

\begin{figure}[ht]
    \centering
    \includegraphics[width = 0.48\textwidth]{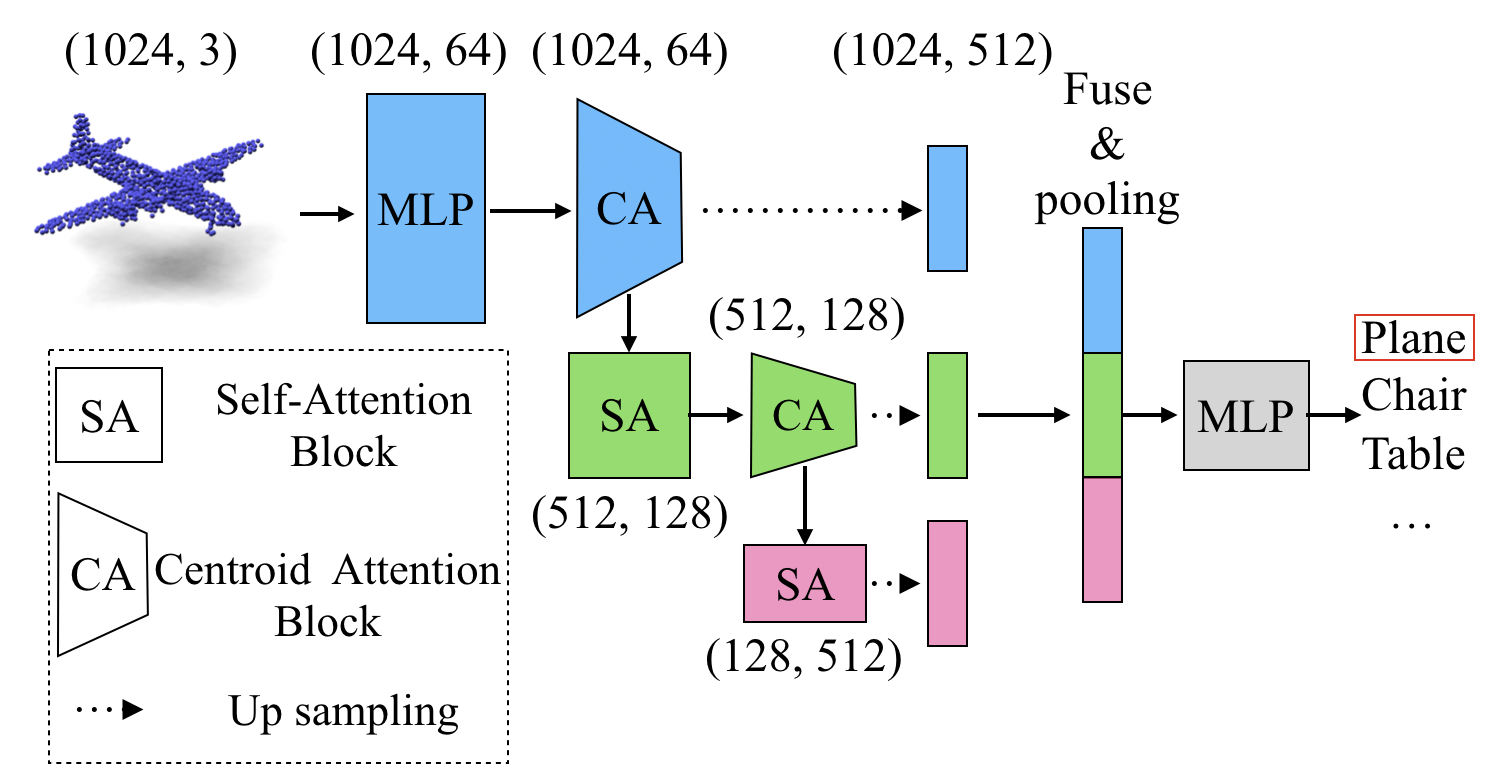}
    \vspace{-0.5em}
    \caption{The architecture of the centroid transformer we used for point cloud classification. ``CA" represents centroid attention  and ``SA"   vanilla self-attention. }
    \label{fig:modelnet40}

\end{figure}

\textbf{Results} \space The results of the experiments is shown in Table.~\ref{tab:nlp}. Centroid transformer with mean-pooling initialization yields the highest score in all three metrics. Moreover, our centroid transformer reduces $~50\%$ MACs against the original transformer on the encoder side. For the same Titan XP GPU, centroid transformer takes less storage space, allowing 38 more samples in a single batch. We also find the initialization scheme to be important. Random sampling scheme results in a centroid transformer that is worse than the baseline.

\subsection{Point Cloud Classification}
We apply our centroid transformer on point cloud classification.
Point cloud data usually has over 1,000 points as input, which is a long sequence and hard to directly train with an original transformer.
By applying our centroid transformer, we can gradually aggregate the thousands points into several abstract clusters, which brings powerful feature extraction ability and saves computational cost.
We compare our method with several baselines including the state-of-the-art attention based method SepNet-W15 \citep{ran2020deeper} on ModelNet40 dataset \citep{wu20153d}. We outperform the existing methods on classification accuracy and consume much less resource comparing with the attention based method SepNet-W15.


\textbf{Data} \space We use ModelNet40 as our benchmark, which has 40 classes with 9843 training shapes and 2468 test shapes. Consider that many of the ModelNet40 classification works use their specifically processed data, for a fair comparison, we use the same data as \citep{qi2017pointnet}, which is an uniformly sampled point cloud data from ModelNet40 raw data without any other specific normalization and preprocess.

\textbf{Model Configuration} \space We design a transformer architecture with 4 attention blocks and a 3-layer MLP classification head as Figure \ref{fig:modelnet40} shows. The first centroid attention block cluster $N$ points to $N/2$ and the second one abstract $N/2$ to $N/8$. We use Farthest Point Sampling (FPS) as our initialization function.  The dimension of the 4 attention locks is 64-128-128-512. We set $K = 40$ in all the KNN masks and the KNN is calculated in feature space in the centroid attention blocks.  After we extract features, we fuse the features in each layer by upsampling their number of points to $N$ using nearest-neighbor sampling . Then we concatenate all the layers' feature together and pass through a fully connected layer. The output feature will pass through a max pooling and a average pooling layer separately. At last, we concatenate them together as the final feature of the input point cloud. The feature will pass a 3-layer MLP to get the final classification score. All the activation functions in the network are LeakyReLU with a negative slope of 0.2. We train the network with Adam optimizer start with $1e^{-3}$ learning rate and decay by $0.7$ every 20 epochs for 250 epochs in total.


\begin{table}[ht]
    \centering
    \scalebox{0.8}{
   \begin{tabular}{l|c|c|c}
        \hline
        Models & Acc & MACs(G) & Params(M) \\
        \hline
        PointNet~\citep{qi2017pointnet} & 89.2 & 0.3 & 0.8 \\
        PointNet++~\citep{qi2017pointnet++} & 91.9 & 7.9 & 12.1  \\
        DGCNN~\citep{wang2019dynamic} & 92.8 & 2.4 & 2.0  \\
        SepNet-W15~\citep{ran2020deeper} & 93.1 & 22.4 & 4.8  \\
        Ours & \textbf{93.2} & 5.4 & 4.1  \\

        \hline
    \end{tabular}}
    \caption{Results on ModelNet40 of our method and  various of baselines. Here MACs denotes multiply-add cumulation and Params means the number of parameters in the model. }
    \label{tab:3d}
\end{table}

\begin{table}[ht]
    \centering
   \begin{tabular}{c|c|c}
        \hline
        Method & Acc & Data/sec  \\
        \hline
        Ours (T=1) & 93.1 & 60 \\
        Ours (T=2) & 93.2 & 55 \\
        Ours (T=3) & 93.2 & 52 \\
        \hline
        Random Sampling & 91.7 & 312 \\
        Farthest Point Sampling & 92.3 & 66 \\
        K-means & 92.3 & 54 \\
        \hline

    \end{tabular}
    \caption{Results on ModelNet40 of centroid attention  when using 
    different numbers of iterations $T$  and different initialization strategies (see   Eq~\eqref{equ:ourmethod}).}
    \label{tab:3d2}

\end{table}

\textbf{Result} \space From Table \ref{tab:3d}, we can see that our model outperforms all the baseline on classification accuracy on ModelNet40. In addition, when comparing with SepNet-W15, our model can achieve a higher accuracy with only 1/4 MACs and fewer parameters.

\textbf{Ablation Study} \space In Table \ref{tab:3d2}, we show the result of our method using different $T$ in centroid attention blocks and compare them with other downsampling strategies. For the different $T$, the performance do not have a big difference. However, larger $T$ means more computation operations, in practice, we choose $T=1$ for a performance-efficiency balance. For the downsampling strategies, we compare our model with Random Sampling, Farthest Point Sampling and K-means. We apply K-means for 3 iterations to guarantee the computation complexity is comparable with our centroid attention. Comparing with farthest point sampling strategy, our speed only slow a little bit while getting a 0.8 improvement in the classification accuracy. Comparing with the random sampling, we have a 1.5 accuracy improvement. In addition, we are 0.9 better than K-means.

\begin{figure}[ht]
    \centering
    \includegraphics[width = 0.45\textwidth]{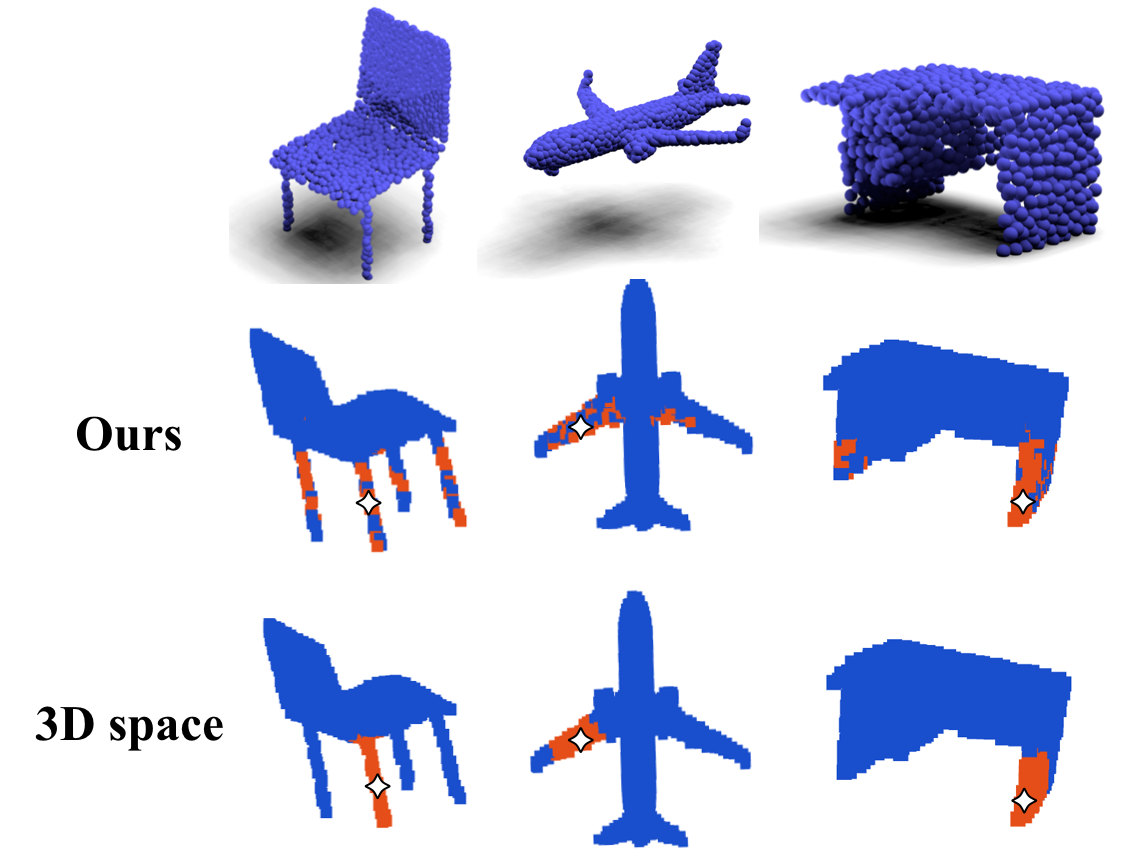}
    \caption{Learning classification on ModelNet40 with  centroid transformer. We
    visualize the K-nearest-neighbours (KNNs) points of some sampled points (white stars). 
    For the KNNs' distance, for ours, we use the L2 distance in feature space learned in the second centroid attention blocks and for 3D space, we use the 3D euclidean distance. Here, $K=40$, The red points indicates the KNNs points of the sampled white point. }
    \label{fig:modelnet40_vis}
    \vspace{-1em}
\end{figure}

\textbf{Visualization} \space We visualize our second centroid attention blocks' feature cluster in Figure \ref{fig:modelnet40_vis}. We plot the sampled point in white star and its K-Nearest-Neighbour (KNNs) in red point. We set $K = 40$ and use L2 distance in KNNs. From Figure \ref{fig:modelnet40_vis}
 we can see that rather than gathering around the sampled point in the 3D space, the KNNs in our features space tend to distribute within the same semantic part of the query part, which indicates our feature captures high-level concept of the shapes.

\begin{figure*}[ht!]
    \centering
    \begin{tabular}{c}
        \includegraphics[width = 0.95\textwidth]{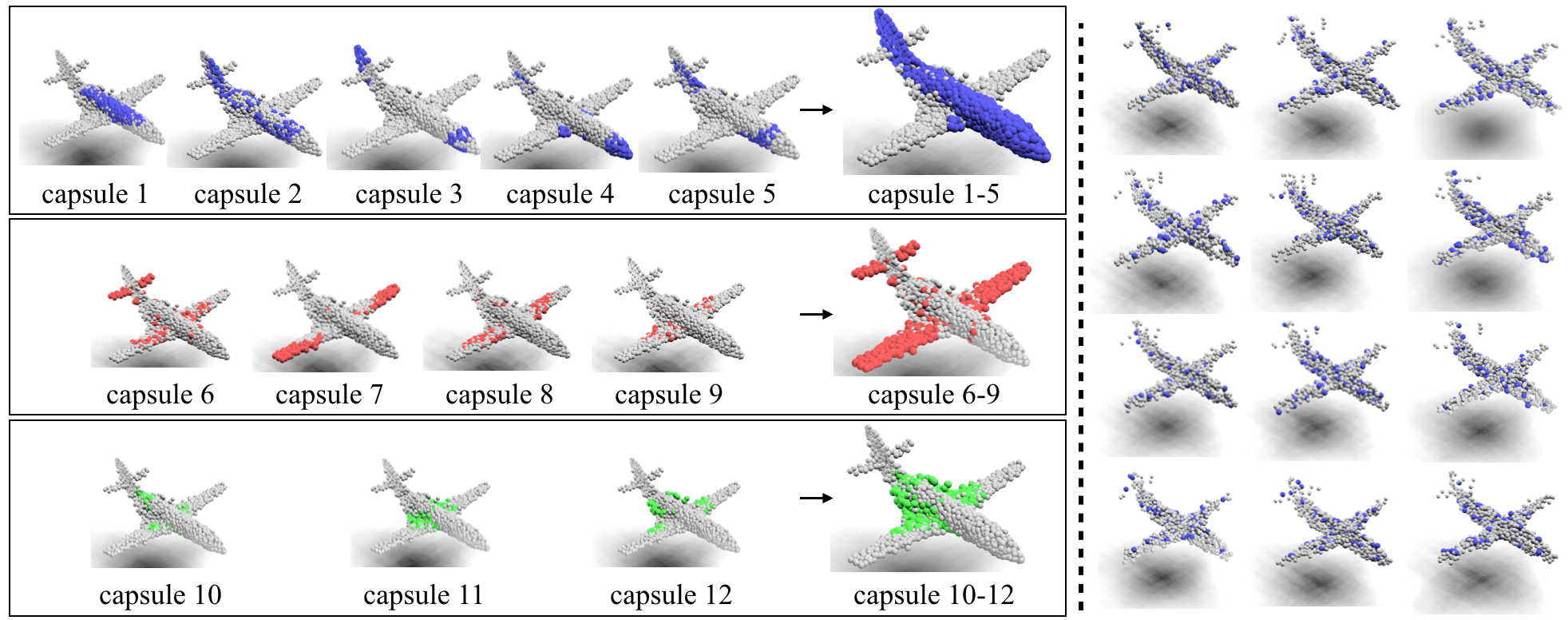}   \\ 
        \hspace{110pt} {\small (a) Capsules learned by our method } \hspace{70pt} 
        {\small (b) Capsules learned by dynamic routing}
    \end{tabular}
    
    \caption{
    Point cloud reconstruction using 3D Capsule Network~\citep{zhao20193d} as backbone, with 12 latent capsules. We decode the 12 latent capsules one by one and highlight the corresponding decoded points with bright colors while keep the other parts gray.
    (a) shows the latent capsules learned by replacing the original dynamic routing module with centroid attention blocks, which  
    can capture semantically meaningful parts of the plane, and are grouped into three clusters that represents plane body, plane front and back wings, and middle engine, respectively. 
    (b) shows the capsules learned by 
    dynamic routing in \citep{zhao20193d}, which distribute randomly and yield no semantic meanings.
    }
    \label{fig:cap_t}

\end{figure*}

\begin{figure*}[ht!]
    \centering
    \begin{tabular}{c}
        \includegraphics[width = 0.98\textwidth]{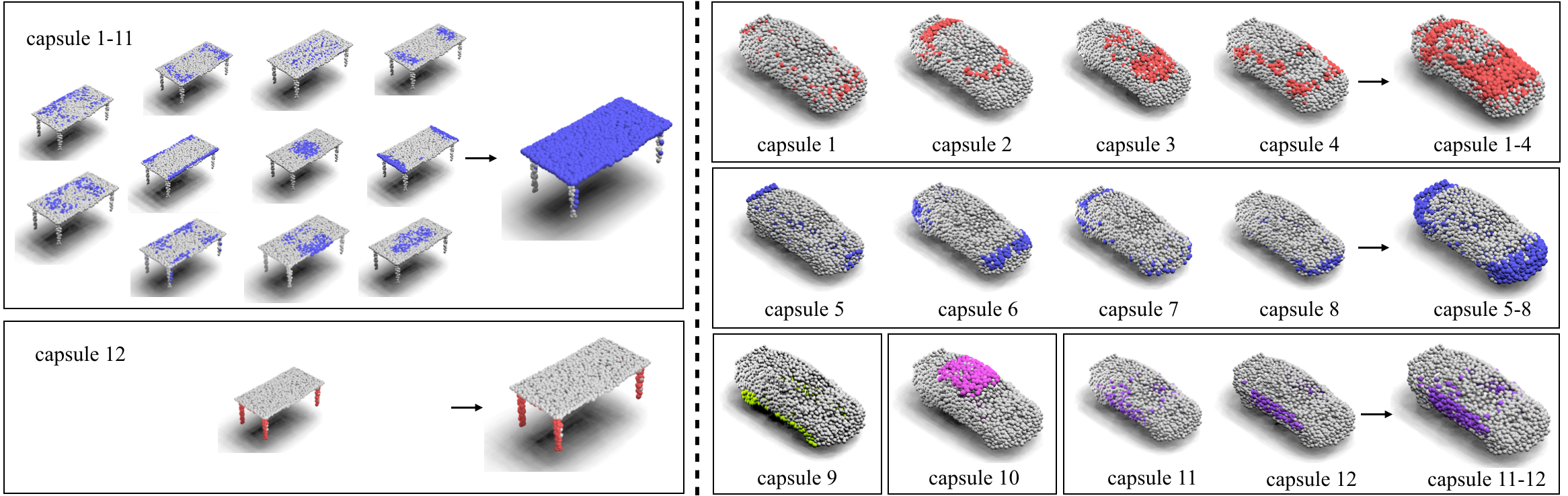}   \\
        \hspace{-40pt} {\small (a) Capsules of our method on a table} \hspace{80pt} 
         {\small (b) Capsules of our method on a car}  
    \end{tabular}
    \caption{More visualization of capsules learned by our method. 
    In both (a) and (b), the capsules map to 
    semantically meaningful parties of the inputs. 
    }
    \label{fig:acap_t}

\end{figure*}

\subsection{Point Cloud Reconstruction}

We further test our centroid attention block on point cloud reconstruction task and make a visualization plot to illustrate our layer's efficiency and abstract ability.
We use ShapeNet part and ShapeNet Core13 \citep{chang2015shapenet} as our dataset and use 3D Capsule Network \citep{zhao20193d} as our backbone.
We replace the Dynamic Routing module in the original capsule network with our centroid attention block to abstract the information from the prime capsules.

\textbf{Experiment Setup} \space We follow the 3D Capsule Network \citep{zhao20193d} settinga and construct an autoencoder to reconstruct the 3D point cloud with 2048 points. We set up two model scales with 12 and 64 latent caspules. In our centroid attention block, we treat the prime capsules as our input sequence and latent capsules as the abstract sequence. We use a linear projection to learn the initialization of the latent capsules. We set $T = 3$ to match number of iterations in Dynamic Routing module.

For the 3D Capsule Network, we setup two network sizes, the small one contains 512 prime capsules with 16 dimensions and 12 latent capsules with 128 dimensions. The larger one contains 1024 prime capsules with 16 dimensions and 64 latent capsules with 64 dimensions. The 64 latent capsuls setting keeps the same with the original setting in 3D capsule network.  We train small setting for 200 epochs using ShapeNet Part dataset and base setting for 300 epochs using ShapeNet Core13 dataset. The other hyperparameter keeps the same as \cite{zhao20193d}.

\textbf{Result} \space From Table \ref{tab:cap} we can see that in the 12 latent capsules setting, our model greatly outperforms the Dynamic Routing module to abstract the latent capsule and leads a Chamfer Distance $1.6 \times 10^{-3}$  comparing with $2.7 \times 10^{-3}$ in baseline.  When comparing with Dynamic Routing in 64 latent capsules setting, which is same as the oringal paper, we still has $10^{-3}$ better Chamfer Distance score comparing with baseline. The results indicate the Dynamic Routing module fails to abstract many prime capsules into a small amount of latent capsules. On the contrast, our centorid transformer can aggregate information no matter the output length is short or long.

Further, our centroid attention block has a much smaller parameter size as well as a larger max batch size per GPU. Tjos means we can process more data in the same time and lead a much faster training speed comparing with the capsule network using Dynamic Routing. This roughly leads a $3 \times $ training speed boosting when training with same amount of GPUs.

\begin{table}[ht]
    \centering
    \scalebox{0.9}{
    \begin{tabular}{c|c|c|c|c}
    \hline
        \# Cap & Method & CD($\times10^{-3}$) & \#Param(M)  & MBS/GPU    
        \\
        \hline

        \multirow{2}{*}{12} & DR & 2.7 & 18.4 & 22  \\
        & Ours & \textbf{1.6} & \textbf{0.6} & \textbf{58} \\

         \hline
        \multirow{2}{*}{64} & DR & 1.5 & 19.8 & 12  \\
        & Ours & \textbf{1.4} & \textbf{5.3} & \textbf{36} \\
        
        \hline

    \end{tabular}}
    \caption{Comparison between the performance of dynamic routing and our centroid attention method
    under different latent capsules number settings (MBS/GPU = Maximum Batch Size per GPU, CD = Chamfer Distance). Here the Chamfer Distance is a metric for evaluating the quality of reconstruction. }
    \label{tab:cap}
\end{table}

\textbf{Visualization} \space We further visualize the decoded points from each latent capsule. To clearly show the semantic meaning, we use 12 latent capsules to plot the reconstruction visualization, in which each capsules can decode into 170 points. Each time, we highlight the decoded points from specific capsules in bright colors and keep the rest of the reconstruction points in gray. In Figure \ref{fig:cap_t}~(a), we show our learned latent capsules can capture semantic meanings and distribute on the specific part of the plane shape. If we group several capsules together, we can further get the whole semantic parts. Figure \ref{fig:cap_t}~(b) shows the decoded points learned by Dynamic Routing. We can find that the reconstruction shape is in a low quality and the highlight points are in a random distribution across the whole shape. That means Dynamic Routing failed to learn a meaningful latent capsules under this setting. We plot two more visualization using our centroid attention blocks to show the result of different shapes with different number of semantic parts. Figure \ref{fig:acap_t} (a) clearly shows that the table are decomposed into the surface and legs. Furthermore, Figure \ref{fig:acap_t} (b) decomposes the car into body, surrounding, wheel, roof and door parts, these illustrate our centroid attention's strong ability in clustering the semantic information.

\subsection{Vision Transformer}
We apply our centroid transformer structure on ImageNet  \citep{deng2009imagenet} classification task using vision transformer to show our capacities among various transformer with large-scale data. We choose DeiT \citep{touvron2020DeiT} as our baseline, which is the state-of-the-art vision classification transformer structure on ImageNet dataset.

\begin{figure}
    \centering
    \includegraphics[width = 0.45\textwidth]{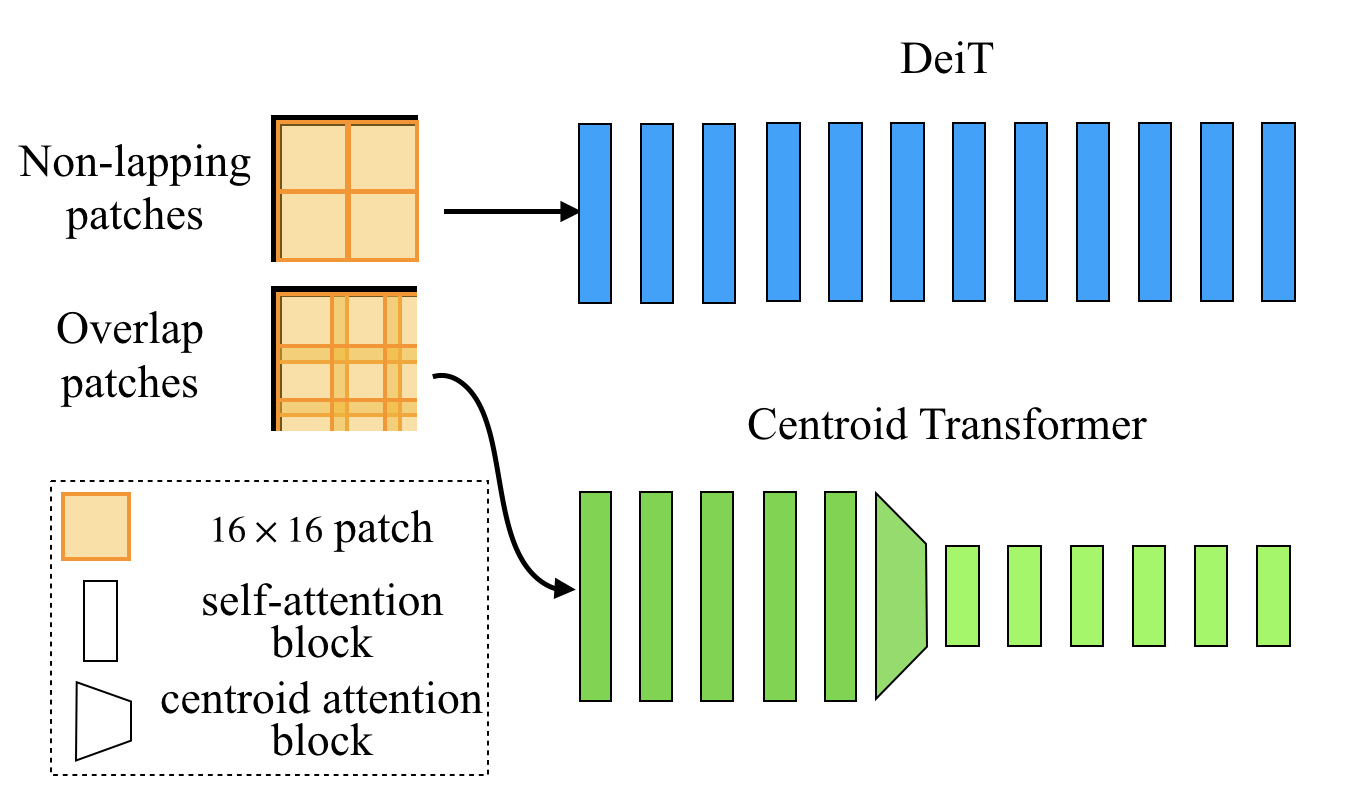}
    \caption{Comparing DeiT and centroid transformer for image inputs. 
    Upper panel: DeiT partitions the input image into non-lapping patches and keeps the same number of patches throughout the layers, which may lose information of the image because the  patches do not overlap. 
    In comparison,  thanks to the ability of down-sampling, the centroid transformer with comparable computational cost can take a larger number of overlapping patches in the early stage and hence capture more information from the input images.
    }
    \label{fig:arch}
 
\end{figure}

By applying centroid attention block in one specific layer, we build up our centroid transformer as Figure \ref{fig:arch} shows. Our transformer abstract the image patches sequence into a shorter one in the middle, which reduces the computational cost comparing with the original DeiT. Further, we allow some overlaps in the first convolution layer to create a longer initial token sequence with richer represent ability.

We compare our centroid transformer with DeiT-tiny and DeiT-small, which are two vision transformers with different model scales. We also apply two efficient transformers, Set Transformer \citep{lee2019set} and Linformer \citep{wang2020linformer} on DeiT as baseline to compare the performance under an energy efficient setting.

\begin{table}[ht]
    \centering
    \scalebox{0.87}{
    \begin{tabular}{c|c|c|c|c}
    \hline
         Models & \#layers & \#tokens & \makecell[c]{embedding \\ dimension} & centroid @  \\
         \hline
         DeiT-tiny & \multirow{4}{*}{12} & 196 & \multirow{4}{*}{192} & - \\
         Ours-1 &  & 196 &  & 6 \\
         Ours-2 &  & 256 &  & 5 \\
         Ours-3 &  & 256 &  & 7 \\
         \hline
         DeiT-small & \multirow{4}{*}{12} & 196 & \multirow{4}{*}{384} & - \\
         Ours-4 &  & 196 &  & 6 \\
         Ours-5 &  & 256 &  & 5 \\
         Ours-6 &  & 256 &  & 7 \\
         \hline
    
    \end{tabular}}
    \caption{Architecture design of different models for image classification tasks on ImageNet, centroid @ means replacing the self-attention block with the centroid attention block at specific layer.}
    \label{tab:arch}
\end{table}

\textbf{Experiment Setup} \space In the overlap setting, we set the first convolution layer with kernel size 16, stride 14, and padding 1. This can create a patch sequence with 256 tokens. In the centroid attention block, we first initialize our shorter sequence by reshaping the N length patch tokens into a $ \sqrt{N} \times \sqrt{N} $ order. We then apply a depth-wise convolution layer with kernel size 3, stride 2, padding 1 to downsample the input tokens into $\sqrt{N}/2 \times \sqrt{N}/2$ and reshape it back to $N / 4$ tokens as the initialization. We set $T=1$ for a better performance-efficiency trade off. The rest of the training setting are the same as DeiT. We setup different model scales. The overall architectures design is listed in Table \ref{tab:arch}. For fast transformer baselines, we set Linformer's project dimension $k=128$ and Set Transformer's induce point set length $m=128$.

\textbf{Result} \space Our results are shown in Tabel \ref{tab:DeiT}. In ``Ours-1'' and ``Ours-4'' setting, by replacing one self-attention blocks into centroid attention block, our model has only 64\% MACs comparing the original DeiT-tiny and DeiT-small while only drop the Top-1 Accuracy by 0.4 and 0.3.  Further, when we allow a longer initial sequence by overlapping the image patchs and adjust the centroid attention blocks to the fifth layer, the centorid block``Ours-2'' and ``Ours-5'' achieves a same and 0.2 higher top-1 with 81\% MACs.
When we further increase the MACs by adjusting the centroid attention block's layer, we can get larger models ``Ours-3'' and ``Ours-6'' with the close MACs comparing with DeiT-tiny and DeiT-small baseline and gains 1.2 and 1.0 performance improvement.

Whatsmore, when we compare centroid transformer with other fast transformers, we outperform set transformer by over 5.0 and linformer by over 2.0 Top-1 accuracy with smaller MACs. Our good performance comparing with these fast transformers further shows our structures' powerful ability and efficiency to handle even this kind of large-scale image classification task.

\begin{table}[ht]
    \centering
    \scalebox{0.9}{
    \begin{tabular}{l|c|c|c}
    \hline
        Method & Top-1 Acc & MACs(G) & Params(M) 
        \\
        \hline
        Set Transformer & 66.1 & 1.1 & 5.9 \\
        Linformer & 69.9 & 1.2 & 6.3 \\
        DeiT-tiny &  72.2 & 1.3 & 5.7 \\
        \hline
        Ours-1 & 71.8 & 0.8 & 5.7 \\
        Ours-2 & 72.2 & 1.0 & 5.7 \\
        Ours-3 & \textbf{73.4} & 1.3 & 5.8 \\
        \hline
        \hline
        Set Transformer & 74.6 & 3.8 & 22.4\\
        Linformer & 77.6 & 4.4 & 22.6  \\
        DeiT-small & 79.9 & 4.7 & 22.1 \\
        \hline
        Ours-4 & 79.6 & 2.9 & 22.1 \\
        Ours-5 & 80.1 & 3.6 & 22.1 \\
        Ours-6 & \textbf{80.9} & 4.7 & 22.3 \\
         \hline

    \end{tabular}
    }
    \caption{Vision Transformer result compared with DeiT. Ours-x indicates different MACs setting of our model. MACs indicates multiply-add cumulation and Params means the number of parameters in the model. }
    \label{tab:DeiT}

\end{table}






%% file: text/related_works.tex


\paragraph{Fast Transformers}
A line of recent works have been developed to approximate self-attention layers to improve over the $O(N^2)$ complexity.
These methods keep the basic design of self-attention, mapping $N$ inputs to $N$ outputs, and is hence different in purpose with our method, which compresses $N$ inputs to a smaller number $M$ of outputs.
Among these works,
Reformer~\citep{kitaev2019reformer} uses Locally Sensitive Hashing to accelerate  attention, yielding a $O(N \log N)$ complexity; LongFormer~\citep{beltagy2020longformer} works by combining sparse global attention with local attention; 
Linear Transformers~\citep{katharopoulos2020transformers}/~Linformer~\citep{wang2020linformer}/~Efficient Attention~\citep{shen2018efficient} achieve linear $O(N)$ complexity by low-rank approximation.
Set Transformer~\citep{lee2019set}
reduces the computation by introducing a set of induced points, which plays a role similar to centroids, only serve as the intermediary layer between $N$ inputs and $N$ outputs.
Another related work is Routing Transformers~\citep{roy2020efficient}, which introduces a sparse routing module based on online k-means, reducing the complexity of attention to $O(n^{1.5})$.

PoWER-BERT~\citep{goyal2020power} also reduces the number of tokens gradually to improve the efficiency of BERT. However, their work fundamentally differs from ours from three aspects: (1) Their work is motivated by the phenomenon of information diffusion in BERT specifically, which may not be the case in other transformers. (2) Their work focus on \emph{select} a subset of tokens from the original sequence, while the emphasis of our work is \emph{summarize} the information into several centroids. This leads to completely distinct structure design. (3) Their scoring function is human-designed. In contrast, we start from clustering algorithm, and derives a novel connection between gradient-based clustering and attention. 


\paragraph{Capsule Networks}
Similar to our method, capsule networks~\citep{hinton2011transforming} are also based on the idea of ``building clustering algorithms into neural networks''.  
Different from our method, which is based on amortizing gradient-based clustering algorithms,
Dynamic routing and EM routing
~\citep{sabour2017dynamic, wang2018optimization, hinton2018matrix} in
capsule networks are based on amortizing EM like algorithms.
However, unlike our method,
 dynamic routing and EM routing are not trainable modules and
 hence not as efficient as our method in extracting data information.
In addition, our method does not need to store the pairwise assignment information like dynamic/EM routing and hence
reduces the runtime space consumption.

\paragraph{Adaptive Down-sampling}
At a high level, our method can be viewed as an attention-like mechanism for adaptively  down-sampling the inputs,
which forms a key building block for deep neural networks in computer vision.
In convolutional neural networks,
various 
techniques have been proposed, including fixed strategies such as
pooling~\citep{simonyan2014very},
strided convolution~\citep{springenberg2014striving}, dilated convolution~\citep{yu2015multi},
learnable down-sampling techniques such as  Local-importance based pooling~\citep{gao2019lip}, deformable convolution~\citep{dai2017deformable}, and trainable Region-of-Interest (ROI) Pooling~\citep{gu2018learning}.
In addition,
\citet{nezhadarya2020adaptive} proposes an adaptive down-sampling layer for point cloud classification. RandLA-Net~\citep{hu2020randla} uses random-sampling to process large-scale point cloud.
Comparing these methods,
which mostly focus on convolutional neural networks (CNNs), our method provides a general and basic adaptive down-sampling strategy for transformers, which is expected to find important applications as the counterparts in CNNs.

\paragraph{Metric Learning for Clustering}
A line works have been developed to
learn domain-specific similarity functions 
to boost
the performance of the clustering algorithms based on the learned metrics~\citep{yang2016joint, yang2017towards, Hsu18_L2C, aljalbout2018clustering, yang2019learning}.
This yields a metric learning task.
Our work is fundamentally different
we use clustering to inspire the design of a new transformer architecture, and our goal is not to actually optimize the clustering quality for a specific problem.

\section{Conclusion}
In this paper, we propose centroid attention, which performs summative reasoning for sequence modeling. Centroid attention takes the original sequence as input, and provides a shorter sequence of centroids that absorbs the information of the input sequence. We use ceontroid attention to construct centroid transformer. By using centroids for later stages, the computational and memory consumption of centroid transformers is significantly reduced against their full-length counterparts. Experiments conducted on text summarization, 3D vision and image processing demonstrates centroid transformers can yield similar / better performance over the original transformers with high efficiency.

%% file: text/appendix_vis.tex
\section{An Energy View of Self-Attention} 
We provide another view to draw connection between attention mechanism and learning to abstract with energy-based models. 
Let's first rewrite the self-attention operation in an energy-view. We start by defining the following energy function on the sequence $\{\vv x_i\}_{i=1}^N$,
\begin{equation}
    E \left(\{\vv x_i\}_{i=1}^N \right) = \sum_{i=1}^N \sum_{j=1}^N \zeta(\bm{x}_i^\top \bm{x}_j),
\end{equation}
where $\zeta(\cdot, \cdot)$ is a pairwise energy function. To find the sequence with the lowest energy, we can perform gradient descent yielding,
\begin{equation}
    \bm{x}_i~\leftarrow~\bm{x}_i -\epsilon \sum_{j=1}^N \nabla_{\bm{x}_i} \zeta(\bm{x}_i^\top \bm{x}_j) \bm{x}_j,~~~\forall i= 1,\ldots, N.
\end{equation}
Properly setting $\zeta(\bm{x}_i^\top \bm{x}_j)$ will recover the single-headed self-attention operation with $v(x_j) = x_j$ 
and a similarity function without the normalization denominator in Eq.~\eqref{eq:softmax}. In this sense, self-attention can be explained as one gradient step towards the sequence with the lowest energy.

The energy function above yields a fully observed pairwise energy function.  The centroid attention can be viewed as corresponding to the energy function of restricted Boltzmann machine (RBM) in which $\{\vv u_j\}_{j=1}^M$ are 
viewed as hidden variables, 
\begin{equation}
    E \left(\{\vv x_i\}_{i=1}^N, \{\vv u_j\}_{j=1}^M \right) = \sum_{i=1}^N\sum_{j=1}^M \zeta(\bm{u}_j^\top \bm{x}_i).
\end{equation}
Here, $\{\vv x_i\}_{i=1}^N$ are the visible variables, and $\{\vv u_j\}_{j=1}^M$ are the hidden variables. Given fixed visible variables, we can also find the hidden variables that minimizes the energy by gradient descent,
\begin{equation}
\label{eq:rbmupdate}
    \bm{u}_j~\leftarrow~\bm{u}_j -\epsilon \sum_{i=1}^N \nabla_{\bm{u}_j} \zeta(\bm{x}_i^\top \bm{u}_j) \bm{x}_i,~~~\forall j= 1,\ldots, M.
\end{equation}
Therefore, centroid attention is like finding the most likely hidden variable for a given observed variable.
Note that the derivation here is different from the one in the main text, because the similarity function is not normalized here. 